# More for Less:

# Compact Convolutional Transformers Enable Robust Medical Image Classification with Limited Data

Andrew Kean Gao

Stanford University

Transformers are very powerful tools for a variety of tasks across domains, from text generation to image captioning. However, transformers require substantial amounts of training data, which is often a challenge in biomedical settings, where high quality labeled data can be challenging or expensive to obtain. This study investigates the efficacy of Compact Convolutional Transformers (CCT) for robust medical image classification with limited data, addressing a key issue faced by conventional Vision Transformers - their requirement for large datasets. A hybrid of transformers and convolutional layers, CCTs demonstrate high accuracy on modestly sized datasets. We employed a benchmark dataset of peripheral blood cell images of eight distinct cell types, each represented by approximately 2,000 low-resolution (28x28x3 pixel) samples. Despite the dataset size being smaller than those typically used with Vision Transformers, we achieved a commendable classification accuracy of 92.49%. The CCT also learned quickly, exceeding 80% validation accuracy after five epochs. Analysis of per-class precision, recall, F1, and ROC showed that performance was strong across cell types. Our findings underscore the robustness of CCTs, indicating their potential as a solution to data scarcity issues prevalent in biomedical imaging. We substantiate the applicability of CCTs in data-constrained areas and encourage further work on CCTs.

## Introduction

The field of bioinformatics leverages the power of computation and big data towards solving problems such as drug discovery, disease diagnosis, vaccine design, and other highly relevant topics [1-11]. Bioinformatics has benefited greatly from the advent of deep learning in recent years. Deep learning has revolutionized the field of biomedical imaging, promoting the development of highly accurate models for tasks such as MRI image classification, X-ray segmentation, and early disease diagnosis [12-14]. Convolutional Neural Networks (CNNs) have been widely used due to their ability to capture local features in images [15-17].

Transformers, developed for natural language processing tasks, have been adapted for image-related tasks, showing promising results [18]. Transformers use the innovation of attention mechanisms to help determine which features are most relevant. Transformers have achieved very astounding results for text and code generation [19-20]. For instance, OpenAI's GPT-3.5 and GPT-4 models have taken the world by storm in late 2022 and 2023. However, a major limitation of transformers is their need for large amounts of training data. This is a significant challenge in the field of medical imaging, where obtaining large, high-quality labeled datasets can be difficult and expensive [21]. Factors that make this difficult include the need for expert annotations (requires significant training and education versus drawing a bounding box over cars, for instance) and the need to protect patient privacy, which can make obtaining data a long process.

To address this challenge, researchers have proposed hybrid models that combine the strengths of CNNs and transformers. One such model is the Compact Convolutional Transformer (CCT), which uses convolutional layers to tokenize images and then transformer layers [22]. CCTs were shown to achieve high accuracy on image classification tasks with modestly sized datasets, making them a promising approach for medical imaging tasks.

The efficacy of CCTs in the context of medical image classification with limited data has not been thoroughly investigated. This is a critical gap, given the scarcity of large, high-quality labeled datasets in medical imaging. This study aims to fill this gap by investigating the performance of CCTs on a benchmark dataset of peripheral blood cell images, with each cell type represented by approximately 2,000 low-resolution samples. The results of this study provide insights into the potential of CCTs as a solution to the data scarcity problem in medical imaging.

## Methods:

The dataset utilized in this study is the BloodMNIST dataset, which is a subset of the MedMNIST benchmark dataset collection [23]. This dataset is composed of microscopic images of eight types of blood cells, which are stained purple. The original source of the data is a data brief by Acevedo et al. titled "A dataset of microscopic peripheral blood cell images for development of automatic recognition systems" [24]. The microscopic images were originally captured at a resolution of $3 \times 360 \times 363$ pixels, but for the purpose of this study, they were center-cropped to $3 \times 200 \times 200$ and subsequently resized to $3 \times 28 \times 28$. The images are squares with three channels (red, green, blue) Because the names of the eight cell types are long, we refer to them by numbers (cell type 0, cell type 1, etc.) and this convention is standardized throughout the figures and tables in this paper.

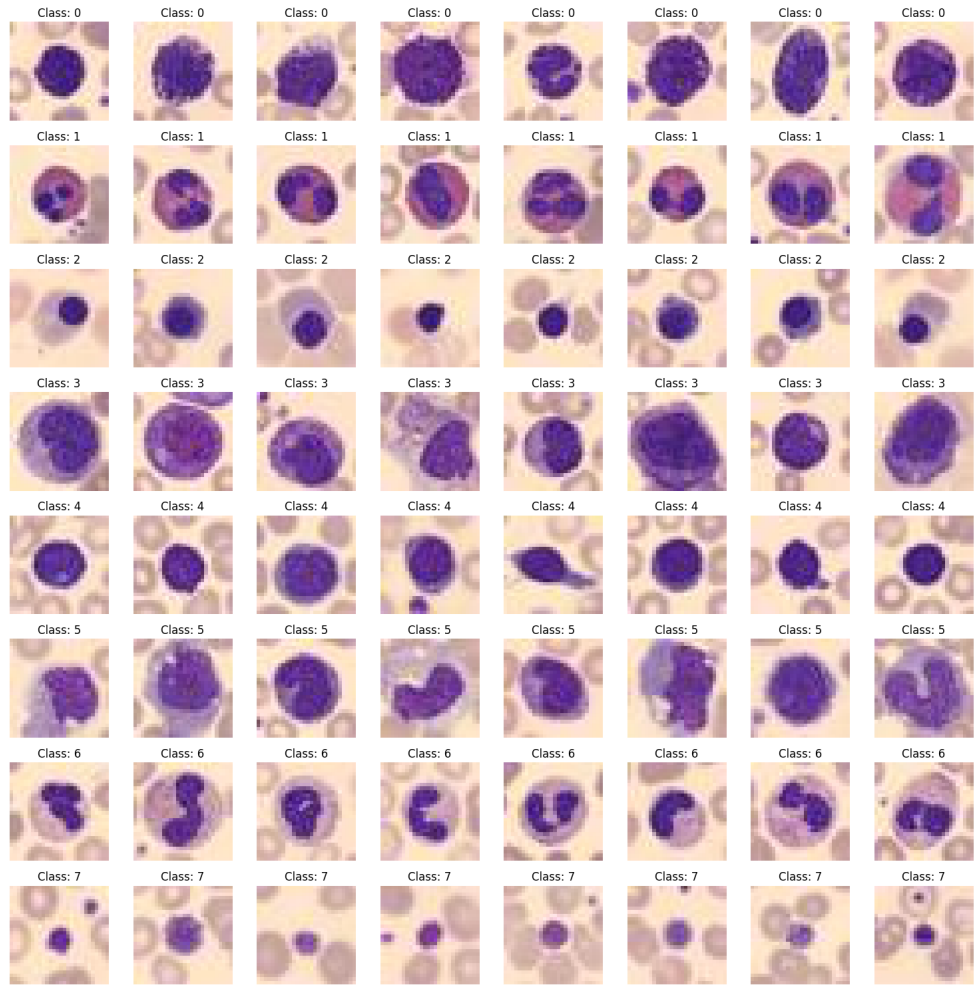

*Figure 1:* Samples for each of the eight cell types. Each row displays eight samples for a given type.

The BloodMNIST dataset contains images of individual normal cells, obtained from individuals who were free of any infection, hematologic disease, or oncologic disease. The individuals were also not under any drug treatment at the time of blood collection. The dataset is organized into eight cell classes: neutrophils, eosinophils, basophils, lymphocytes, monocytes, immature granulocytes, erythroblasts, and platelets or thrombocytes. The cells were stained purple using the May Grünwald-Giemsa stain in the Sysmex SP1000i machine. The images of the cells were captured using the CellaVision DM96 analyzer at the Core Laboratory at the Hospital Clinic of Barcelona in Spain. The labels for each cell image were decided by expert pathologists.

| Cell Type | Samples | % of total dataset |
|---|---|---|
| Neutrophils | 3,329 | 19.48 |
| Eosinophils | 3,117 | 18.24 |
| Basophils | 1,218 | 7.13 |
| Lymphocytes | 1,214 | 7.1 |
| Monocytes | 1,420 | 8.31 |
| Immature Granulocytes | 2,895 | 16.94 |
| Erythroblasts | 1,551 | 9.07 |
| Platelets (thrombocytes) | 2,348 | 13.74 |
| **Total** | **17,092** | 100 |

*Table 1:* Composition of the BloodMNIST dataset by type.

The files for the BloodMNIST dataset are saved in NumPy npz format, named as "bloodmnist.npz" for the 2D dataset. There were 17,092 samples. Of this 17,092, 11,959 were in the training set, 1,712 were in the validation set, and 3,421 were in the test set. We did not use the samples in the validation dataset from bloodMNIST to see if having even fewer samples would suffice for training our CCT.

We used the medmnist library to access and load the BloodMNIST dataset. We split the dataset into training and test (90% and 10%, respectively). The images were converted to Numpy arrays and normalized to the range of 0-1. In each batch during training, images were randomly cropped and flipped to combat overfitting. The labels were one-hot encoded using Keras's to_categorical function. We defined the Compact Convolutional Transformer's architecture using the Keras functional API, based on the implementation provided in the Keras documentation (based on Hassani et al's paper) [25]. We use the approach of an all-convolution mini-network to create the image patches. The CCTTokenizer layer performs tokenization with convolutional layers and max pooling layers, splitting each training image into smaller non-overlapping squares. Next, it implements positional embedding so the spatial relation information is not lost. This is the same technique used in language transformer models like GPT4 so that the order of words is not lost.

The tokenized image patches are passed through eight transformer layers. Each layer contains a multi-head self attention mechanism with four heads which is followed by a feed-forward neural network. We apply stochastic depth regularization and layer normalization to reduce overfitting. Stochastic depth regularization will randomly drop a set of layers during training which helps combat overfitting. The output from the transformer layers is then passed through a layer normalization and a dense layer in order to calculate attention weights. The weighted representation of the patches is sent through another dense layer to calculate the output logits. The model was compiled using the AdamW optimizer with a learning rate of 0.0018 and a weight decay of 0.00012. The batch size was set to 64 and the number of epochs was set to 75. The loss function we used was categorical cross-entropy loss with label smoothing. We used callbacks to save the best performing model checkpoints and apply it to the testing set at the end of training. We evaluated model performance using top-1 and top-2 accuracy.

We plotted training and validation loss and accuracy curves to visualize the training progress and identify potential overfitting. We also calculated and plotted multi-class receiver operating characteristic (ROC) curves to further assess our model's performance. We then generated a confusion matrix to contrast predicted versus true labels in our test set. Next, we printed the default classification report to analyze the precision, recall, F-1 score, and support for each class. The model was saved to an HDF5 file and will be made available.

## Results:

A total of 11,959 samples were used to train the model for 75 epochs, with 10% reserved for the validation set. Validation loss decreased sharply in the first five epochs and quickly leveled off. Validation loss was somewhat erratic, which could be partially attributed to the smaller size of the validation set (approximately 1,200 images). After around 50 epochs, loss and accuracy did not improve. Using checkpoints, we restored the best performing model weights and used them to predict on the test set of 3,421 unseen sample images. The classification accuracy on the test set was 92.49% (3,421 samples).

We examined the precision, recall, and F1 score for each of the eight cell types. The F1 score combines precision and recall. Cell type 5 had a precision of 0.91 and a recall of 0.73, significantly weaker than other cell types. This indicates that the model had difficulty correctly identifying cells of type 3. On the other hand, it was very good at identifying cells of type 7 (Precision: 0.99, Recall: 0.99, F1: 0.99).

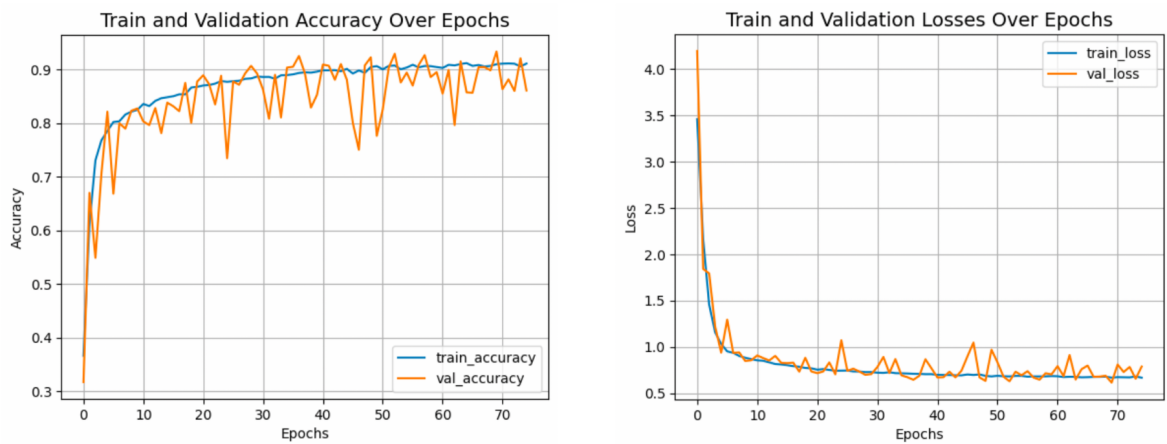

*Figure 2:* (Left) Training and validation loss decreases rapidly during the first epochs. (Right) Training and validation accuracy increases rapidly during the first epochs.

| Class | Precision | Recall | F1-Score | Support |
|---|---|---|---|---|
| 0 | 0.9 | 0.86 | 0.88 | 244 |
| 1 | 0.99 | 0.98 | 0.99 | 624 |
| 2 | 0.95 | 0.87 | 0.91 | 311 |
| 3 | 0.8 | 0.91 | 0.85 | 579 |
| 4 | 0.95 | 0.89 | 0.92 | 243 |
| 5 | 0.91 | **0.73** | **0.81** | 284 |
| 6 | 0.93 | 0.98 | 0.95 | 666 |
| 7 | 0.99 | 0.99 | 0.99 | 470 |
| Accuracy | | | **0.9249** | 3421 |
| Macro Avg | 0.93 | 0.9 | 0.91 | 3421 |
| Weighted Avg | 0.93 | 0.92 | 0.92 | 3421 |

*Table 2*: Classification report showing the precision, recall, F1 score, and support (number of samples) for each cell type. The overall accuracy was 92.49%.

We calculated the Receiver Operating Characteristic curve for each of the eight classes using One Versus Rest strategy. This showed robust classification performance for all eight types, although some types were more accurately classified than others. The lowest AUC score was 0.9833 (cell type 5) while the highest was 0.9999 (cell type 7). The micro-average ROC curve had an AUC of 0.9935. We also generated a confusion matrix to show correct and incorrect predictions on the test set by cell type. This revealed that the model often misclassified cells of type 3, which was corroborated by the lower AUC score for cell type 3.

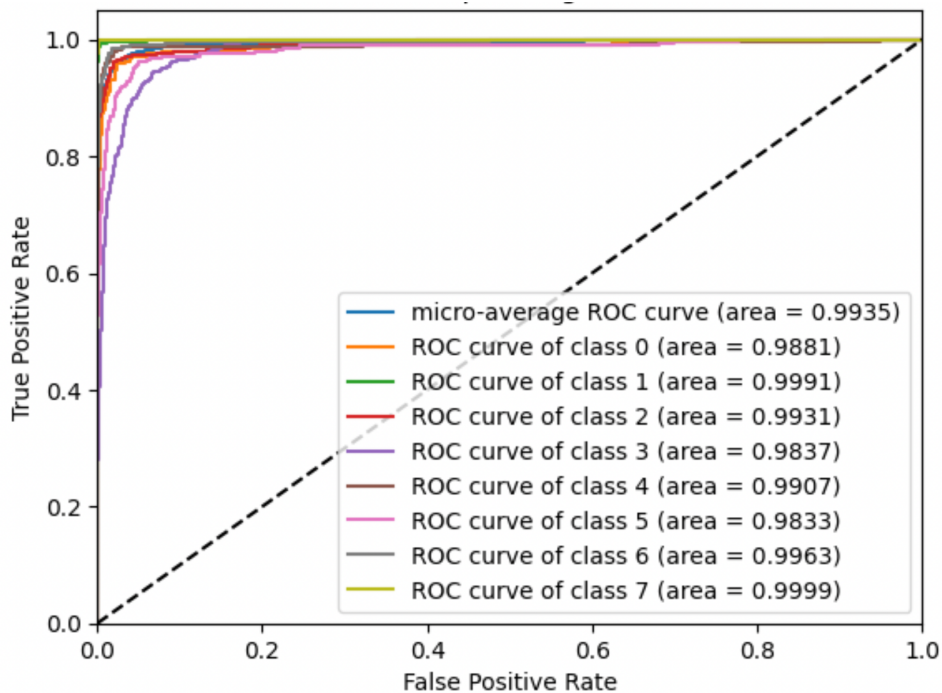

*Figure 3:* Multiclass ROC curve shows highly robust but varied performance across cell types.

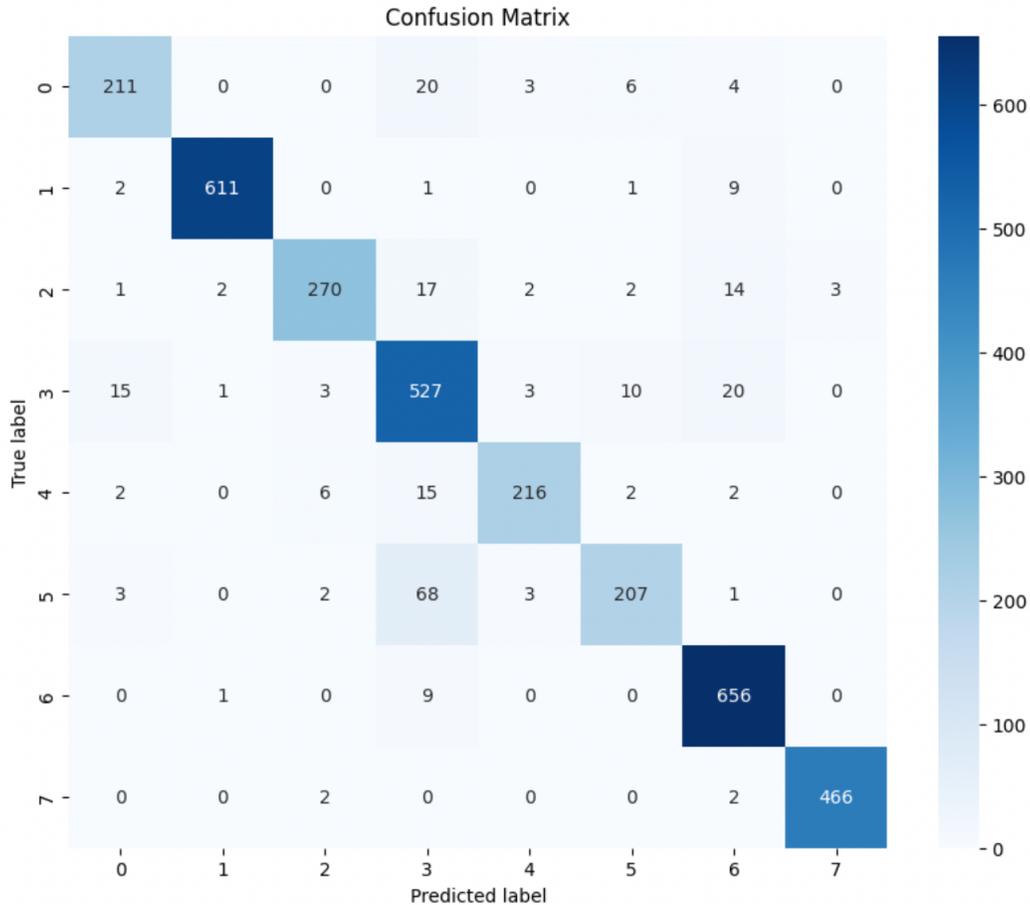

*Figure 4:* Confusion matrix displays true versus predicted labels for each of the eight cell types. Squares on the diagonal represent correct classifications (true = predicted). A darker color indicates more samples. For instance, 656 instances of cell type 6 were correctly classified as such.

# Conclusion:

One of the challenges faced by Vision Transformers is their need for large amounts of data for training, a factor that has fueled the practice of pretraining on very large datasets. The need for "big data" can be a significant impediment, especially in niche fields such as biomedical imaging where labeled data can be scarce or expensive to collect. The results of our study demonstrate that a hybrid of transformers and convolution layers, Compact Convolutional Transformers (CCT), can achieve high accuracy in image classification tasks despite limited training data.

In this research, we applied CCT in a setting characterized by a limited dataset - a common scenario in biomedical imaging. Specifically, we used a relatively modest benchmark dataset composed of 17,092 peripheral blood cell images, covering eight distinct cell types. The dataset had around two thousand low-resolution (28x28x3) samples for each of eight cell types on average. Despite the size of the dataset being smaller than what would normally be preferred for Vision Transformers, we achieved a noteworthy classification accuracy of 92%.

Our findings demonstrate the robustness of Compact Convolutional Transformers, presenting an effective solution to the perennial problem of data scarcity in biomedical imaging. Additionally, the ability to achieve high accuracy using low-resolution images suggests that CCTs may be robust to variations in image quality, which is another common challenge in medical imaging.

In conclusion, our results substantiate the applicability and effectiveness of Compact Convolutional Transformers in contexts where data is limited. This opens up a wide array of possibilities for the application of this technology in biomedical imaging and other fields where high-quality, large volume data may be hard to come by. Compact Convolutional Transformers are a significant step towards democratizing the use of transformers in niche, data-constrained areas.